\documentclass{article}
\PassOptionsToPackage{sort&compress,numbers}{natbib}
\usepackage[final]{neurips_2019_ml4ad}

\usepackage[utf8]{inputenc} % allow utf-8 input
\usepackage[T1]{fontenc}    % use 8-bit T1 fonts
\usepackage{hyperref}       % hyperlinks
\usepackage{url}            % simple URL typesetting
\usepackage{booktabs}       % professional-quality Fig.s
\usepackage{amsfonts}       % blackboard math symbols
\usepackage{nicefrac}       % compact symbols for 1/2, etc.
\usepackage{microtype}      % microtypography
\usepackage{graphicx}
\usepackage[font=scriptsize]{subcaption}
\usepackage{wrapfig,lipsum,booktabs}
\usepackage[font=small]{caption}
\graphicspath{ {images/} }
\usepackage{xcolor}
\title{On the Role of Receptive Field in \\Unsupervised Sim-to-Real Image Translation}
\author{%
  Nikita Jaipuria\textsuperscript{1},~ Shubh Gupta\textsuperscript{2}\thanks{The author contributed to this work during his time as an intern with Ford Motor Company.},~ Praveen Narayanan\textsuperscript{1},~ Vidya N. Murali\textsuperscript{1}\\
  \textsuperscript{1}Ford Motor Company,~ \textsuperscript{2}Stanford University\\
  \texttt{njaipuri@ford.com},~ \texttt{shubhgup@stanford.edu},~ \texttt{pnaray11@ford.com},~ \texttt{vnariyam@ford.com}
}

\begin{document}
\maketitle

\begin{abstract}
%\todo{We investigate the role of the \texttt{D}'s receptive field in unsupervised image-to-image translation with mismatched data, using Generative Adversarial Networks (GANs). Extensive experimentation with the \texttt{D} architecture of a state-of-the-art coupled Variational AutoEncoder (VAE) - GAN model~\cite{liu2017unsupervised} shows that reducing the receptive field of the \texttt{D} is directly correlated to improved semantic content retention in sim-to-real translations.} 
Generative Adversarial Networks (GANs) are now widely used for photo-realistic image synthesis. In applications where a simulated image needs to be translated into a realistic image (sim-to-real), GANs trained on unpaired data from the two domains are susceptible to failure in semantic content retention as the image is translated from one domain to the other. This failure mode is more pronounced in cases where the real data lacks content diversity, resulting in a content \emph{mismatch} between the two domains - a situation often encountered in real-world deployment. In this paper, we investigate the role of the discriminator's receptive field in GANs for unsupervised image-to-image translation with mismatched data, and study its effect on semantic content retention. Experiments with the discriminator architecture of a state-of-the-art coupled Variational Auto-Encoder (VAE) - GAN model on diverse, mismatched datasets show that the discriminator receptive field is directly correlated with semantic content discrepancy of the generated image.
\end{abstract}

%%%%%% INTRODUCTION %%%%%%
\section{Introduction}
\label{sec:intro}
Advanced Driver Assistance Systems (ADAS) and self-driving cars utilize Deep Neural Networks (DNNs) for a variety of perception tasks. The main bottleneck in training and validating DNNs is the collection and annotation of large quantities of diverse data. As an alternative, gaming-engine based simulations can quickly generate large quantities of diverse synthetic data with accurate ground truth. However, models trained on synthetic data often fail to generalize to the real-world because of lack of realism in synthetic data.

Prior works on GANs~\cite{goodfellow2014generative} and VAEs~\cite{kingma2013auto} have shown promising results in photo-realistic image synthesis, in both supervised and unsupervised settings. In the unsupervised setting, the training data consists of \emph{unpaired} images in the simulated and real domains. Here, unpaired refers to the: \emph{(i)} absence of one-to-one correspondence between the simulated and real images used for training; and \emph{(ii)} absence of any form of annotation, such as semantic segmentation masks or edge masks. We are also interested in the unsupervised setting, since it is less restrictive in terms of training data requirements. Additionally, we assume lack of content diversity in the real training data, which causes a content mismatch between the two datasets. For instance, consider an object detection task on trailer images where the real dataset has images of trailer type A and the simulated dataset has images of other trailer types to add diversity in trailer types to the full dataset. While such a problem setting is more challenging, it is also more practical, intentional and often encountered in real-world deployments.

\begin{figure}[h]
    \begin{subfigure}[t]{0.33\linewidth}
        \includegraphics[trim={2cm .5cm 3.5cm 0},clip,height=.49\linewidth]{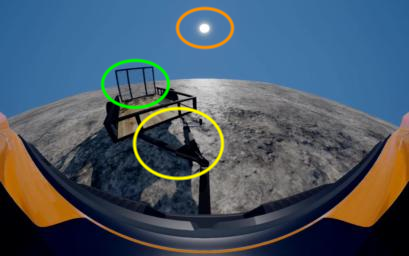}
        \includegraphics[trim={2cm .5cm 4.5cm 0},clip,height=.49\linewidth]{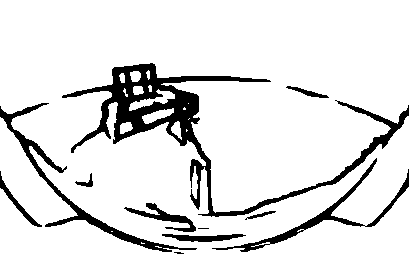}
        \caption{\centering Input Simulated Image}
        \label{fig:unit11}
    \end{subfigure}
    \begin{subfigure}[t]{0.33\linewidth}
        \includegraphics[trim={2cm .5cm 3.5cm 0},clip,height=.49\linewidth]{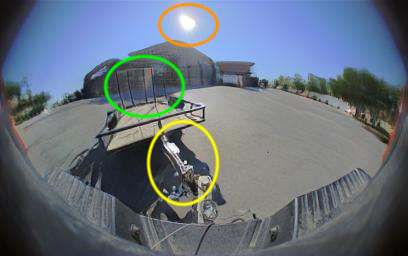}
        \includegraphics[trim={2cm .5cm 4.5cm 0},clip,height=.49\linewidth]{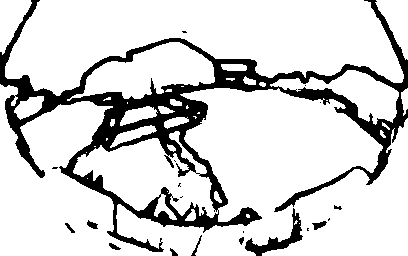}
        \caption{\centering Sim-to-real with UNIT ($46\times 46$)}
        \label{fig:unit12}
    \end{subfigure}
    \begin{subfigure}[t]{0.33\linewidth}
        \includegraphics[trim={2cm .5cm 3.5cm 0},clip,height=.49\linewidth]{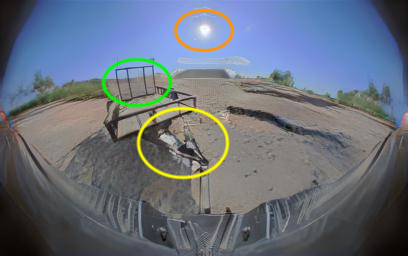}
        \includegraphics[trim={2cm .5cm 4.5cm 0},clip,height=.49\linewidth]{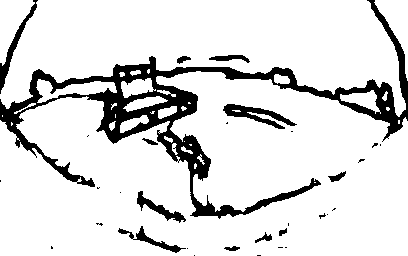}
        \caption{\centering Sim-to-real with Model 4 ($16 \times 16$)}
        \label{fig:unit13}
    \end{subfigure}
    \caption{The effect of reducing \texttt{D}'s receptive field on unsupervised sim-to-real translation with mismatched data: \textbf{(a)} Simulated image of an A-frame trailer (left) and its Holistically-nested Edge Detections (HED) (right); \textbf{(b)} Sim-to-real translation on (a) with UNIT (left) and its HED (right). Note the shift in position of sun (orange), change in trailer coupler shape from A-frame to straight (yellow) and change in number of vertical bars on trailer (green); \textbf{(c)} Sim-to-real translation on (a) with modified UNIT, where the original \texttt{D} was replaced with one with a reduced receptive field (left) and its HED (right). Note the improved semantic content retention between (a) and (c) as compared to that between (a) and (b).\label{fig:unit}}
\end{figure}

In the unsupervised setting, prior works such as Unsupervised Image-to-Image Translation (UNIT)~\cite{liu2017unsupervised}, combine the VAE reconstruction error, the adversarial GAN loss, cycle consistency constraints~\cite{zhu2017unpaired} and perceptual losses~\cite{johnson2016perceptual} to show promising results. However, when such unsupervised methods are applied to datasets that are heavily mismatched in terms of semantic content, these constraints are not enough to ensure retention of low-level scene semantics. Fig.~\ref{fig:unit11} and Fig.~\ref{fig:unit12} show sim-to-real results with UNIT trained on unpaired simulated and real trailer images. Here all simulated images are of trailers with \emph{A-frame} couplers, while all real images are of trailers with \emph{straight} couplers. Note that while high level content is preserved during sim-to-real translation from Fig.~\ref{fig:unit11} to Fig.~\ref{fig:unit12}, lower-level details, such as, trailer coupler shape 
(A-frame vs. straight), structure (number of vertical bars) and sun position, are not preserved. This renders ground truth labels from simulation unusable in any downstream perception task.

We argue that these discrepancies are due to the large content mismatch between the simulated and real datasets. GANs simultaneously 
train adversarial networks – a generator (\texttt{G}) and a discriminator (\texttt{D}) – that compete to generate realistic imagery 
and distinguish between synthetic and real imagery, respectively. Since all the real images that \texttt{D} sees during training have 
trailers with straight couplers, \texttt{G} quickly converges to a point where it starts replacing A-frame couplers with straight couplers. 
To address this issue, we introduce a simple yet surprisingly powerful notion - \textit{what if we reduce the receptive field of \texttt{D} 
such that it can only see parts of, but never the full trailer coupler?} Fig.~\ref{fig:unit13} shows sim-to-real results with a modified 
UNIT architecture, where the original \texttt{D} (with a receptive field of $46\times 46$) is replaced with a modified \texttt{D} (with a 
receptive field of $16\times 16$). Observe the improved retention in trailer coupler shape, number of vertical bars on trailer and sun position. 
This is primarily because the modified \texttt{D} penalizes structure at the scale of image patches that are much smaller than the size of the
trailer coupler itself. The rest of the paper describes experimentation with the network architecture of \texttt{D}, which establish the hypothesis that a reduction in \texttt{D}'s receptive field is directly correlated to improved semantic content retention during sim-to-real translation with GANs.

\vspace{-0.15in}

\begin{figure}[h]
\centering
\begin{minipage}[b]{0.425\textwidth}
\centering
\begin{tabular}{|c|c|c|} 
  \hline
  \tiny{\textbf{Model}} & \tiny{\textbf{Discriminator Architecture}} & \tiny{\textbf{Receptive Field}} \\ [0.5ex] 
  \hline\hline
  \tiny UNIT & \tiny $A \to A \to A \to A \to C$ & \tiny $46 \times 46$ \\ 
  \tiny 1 & \tiny $D \to D \to A \to C$ & \tiny $40\times 40$ \\
  \tiny 2 & \tiny $A \to A \to A \to C$ & \tiny $22\times 22$ \\
  \tiny 3 & \tiny $A \to B \to B \to B \to C$ & \tiny $22\times 22$ \\
  \tiny 4 & \tiny $A \to B \to B \to C$ & \tiny $16\times 16$ \\
  \tiny 5 & \tiny $A \to A \to C$ & \tiny $10\times 10$ \\
  \tiny 6 & \tiny $A \to B \to C$ & \tiny $10\times 10$ \\
  \tiny 7 & \tiny $B \to B \to B \to C$ & \tiny $10\times 10$ \\
  \tiny 8 & \tiny $B \to B \to C$ & \tiny $7\times 7$ \\ [1ex] 
  \hline
\end{tabular}
\centering
\caption{List of UNIT and its modifications. $A, B, C$ and $D$ denote 2D convolutional layers with kernel sizes of $4, 4, 1, 4$ and strides of $2, 1, 1, 3$ respectively.}
\label{fig:dismode}
\end{minipage}
\qquad
\begin{minipage}[b]{0.5\textwidth}
  \centering
  \includegraphics[width=0.85\linewidth]{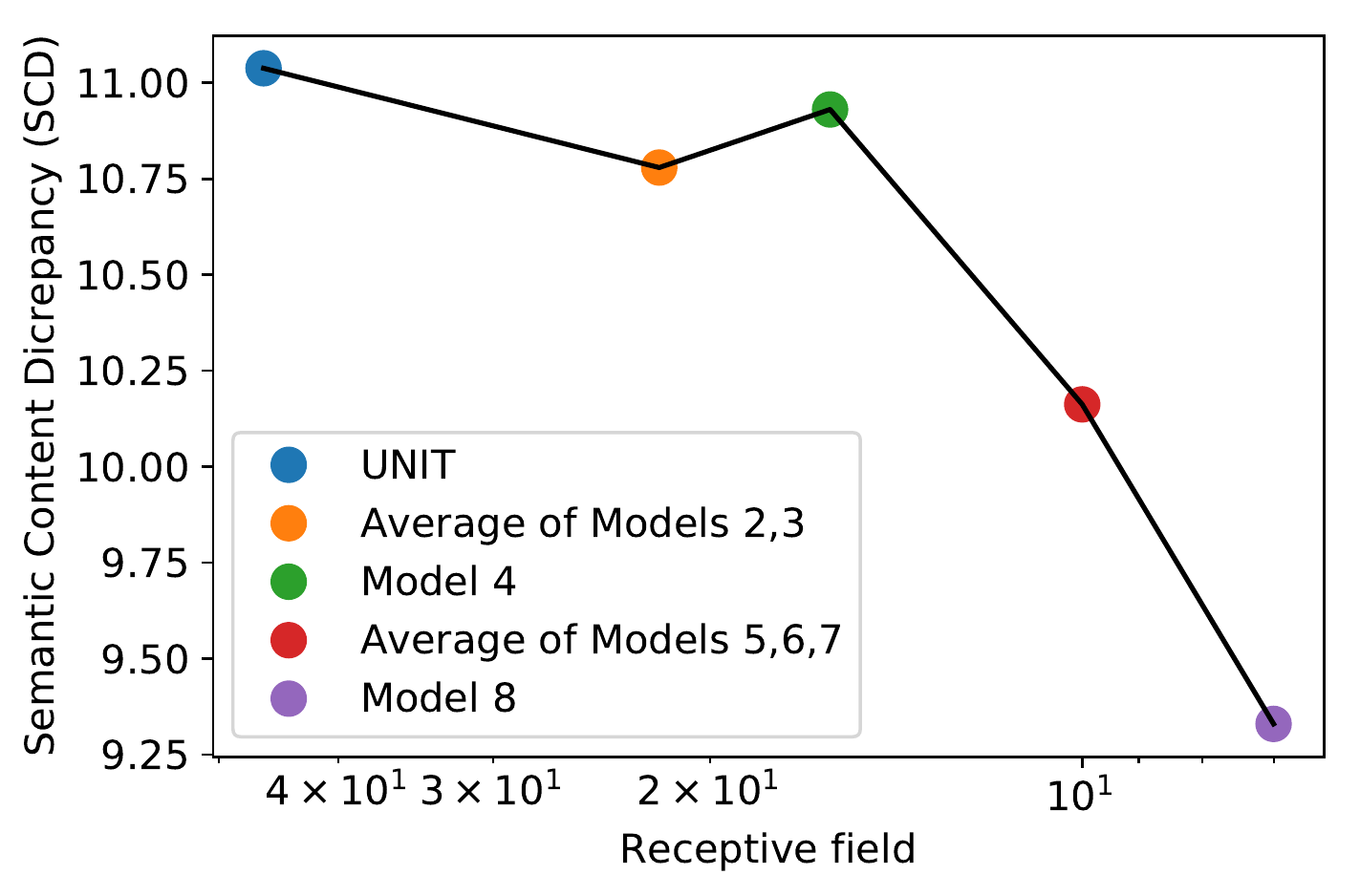}
  \caption{Plot of SCD vs. receptive field. SCD values for models with the same receptive field are averaged.}
  \label{fig:mhd}
  \vspace{.25cm}
  \end{minipage}
\end{figure}

%%%%%% PRELIMINARIES %%%%%%
\section{Preliminaries}
\label{sec:prelim}
\textbf{PatchGAN:} PatchGAN~\cite{isola2017image} has a convolutional \texttt{D} that penalizes structure at the scale of image patches, while simultaneously capturing local style statistics. As opposed to classifying the full image as real or fake, a PatchGAN \texttt{D} classifies $N\times N$ image patches as real or fake. The final output for an image is the average of the patch-wise real/fake predictions. \citet{isola2017image} showed high quality results even with patch sizes ($N$) that were much smaller than the image size. Given mismatched data, a \texttt{D} that classifies large image patches as real or fake prevents generation of semantic content (e.g.\ trailer shapes) not present in the real domain (e.g.\ A-frame). To this end, limiting \texttt{D}'s \emph{field-of-view} to smaller image patches, wherein the semantic entities present are only visible in part, can help preserve content during translations.

\textbf{Unsupervised Image-to-Image Translation (UNIT):} The UNIT architecture is a coupled VAE-GAN architecture, which makes use of a shared latent space assumption~\cite{liu2017unsupervised} to learns sim-to-real and real-to-sim models simultaneously. Thus, UNIT is comprised of two \texttt{D}s, one for each domain. Each \texttt{D} is a 5-layer PatchGAN. The first 4 convolutional layers have a kernel size of 4 and stride of 2, while the last layer has a kernel size of 1 and stride of 1. The effective receptive field ($N$) of each of the two \texttt{D}s is $46$. UNIT is used as the baseline architecture in this work.

\section{Modifications to UNIT} 
We explored two different ways of reducing \texttt{D}'s receptive field in UNIT: (i) reducing the number of layers; and (ii) reducing stride; and studied the effect of each on semantic content retention. Architectural variants of \texttt{D} with a kernel size other than the original size of $4\times 4$~\cite{liu2017unsupervised} caused a drastic drop in visual quality of the sim-to-real translations. Thus, kernel size was not varied in our experiments. Models 2-8 in Fig.~\ref{fig:dismode} are the architectural variants of \texttt{D} that were used to investigate the role of \texttt{D}'s receptive field in sim-to-real image translation.

Since reducing the number of layers in \texttt{D} also reduces its representation capacity, improved sim-to-real translations with Models 2 and 4-8 in Fig.~\ref{fig:dismode}) could also be attributed to reduced overfitting, as opposed to a reduction in \texttt{D}'s receptive field. To investigate the role of overfitting, an additional experiment was conducted in which \texttt{D}'s receptive field was kept similar to that in UNIT, while reducing the number of layers from 5 to 4 (refer Model 1 in Fig.~\ref{fig:dismode} with a discriminator receptive field of $40\times 40$ vs. UNIT with a discriminator receptive field of $46\times 46$). For this purpose, no further experimentation was done as it is not possible to achieve a receptive field similar to that of $46 \times 46$ with fewer layers (i.e. $< 4$) for a kernel size of $4\times 4$.

%%%%%% EXPERIMENTS %%%%%%
% \section{Dataset}
% \label{sec:drpd}

%%%%%% RESULTS %%%%%%
\section{Experiments}
\label{sec:results}
\begin{figure}[h]
    \centering
    \begin{subfigure}[t]{0.16\linewidth}
        \includegraphics[trim={2cm .5cm 3.5cm 0},clip,height=.95\linewidth]{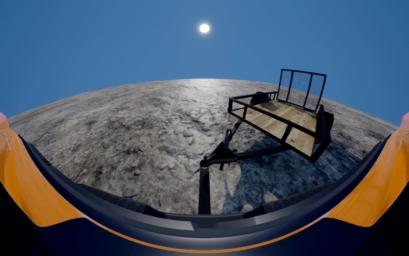}
        \label{fig:result11}
    \end{subfigure}
    \begin{subfigure}[t]{0.16\linewidth}
        \includegraphics[trim={2cm .5cm 3.5cm 0},clip,height=.95\linewidth]{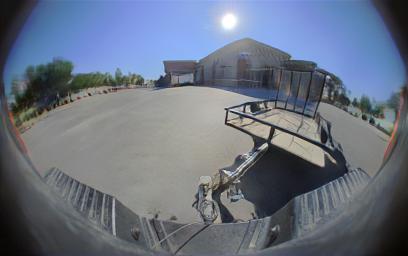}
        \label{fig:result12}
    \end{subfigure}
    \begin{subfigure}[t]{0.16\linewidth}
        \includegraphics[trim={2cm .5cm 3.5cm 0},clip,height=.95\linewidth]{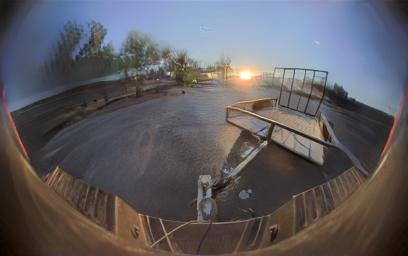}
        \label{fig:result13}
    \end{subfigure}
    \begin{subfigure}[t]{0.16\linewidth}
        \includegraphics[trim={2cm .5cm 3.5cm 0},clip,height=.95\linewidth]{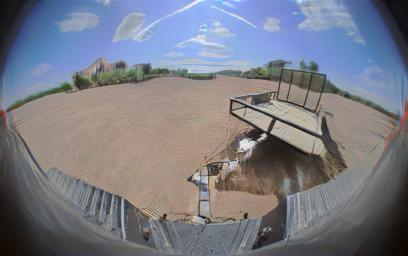}
        \label{fig:result14}
    \end{subfigure}
    \begin{subfigure}[t]{0.16\linewidth}
        \includegraphics[trim={2cm .5cm 3.5cm 0},clip,height=.95\linewidth]{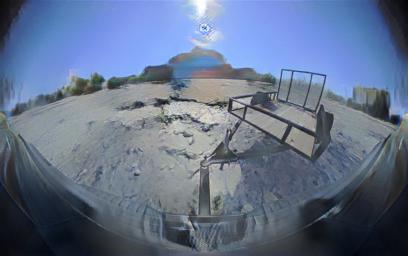}
        \label{fig:result15}
    \end{subfigure}
    \vspace{-0.3cm}
    \begin{subfigure}[t]{0.16\linewidth}
        \includegraphics[trim={2cm .5cm 3.5cm 0},clip,height=.95\linewidth]{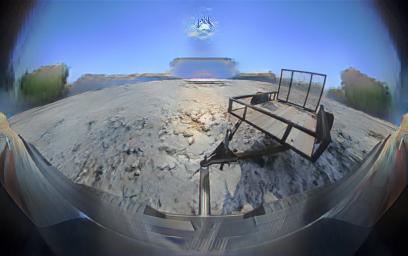}
        \label{fig:result16}
    \end{subfigure}
    \vspace{-0.3cm}
    \begin{subfigure}[t]{0.16\linewidth}
        \includegraphics[trim={2cm .5cm 3.5cm 0},clip,height=.95\linewidth]{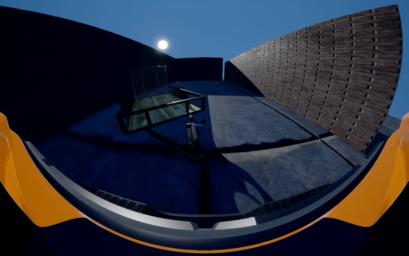}
        \label{fig:result17}
    \end{subfigure}
    \begin{subfigure}[t]{0.16\linewidth}
        \includegraphics[trim={2cm .5cm 3.5cm 0},clip,height=.95\linewidth]{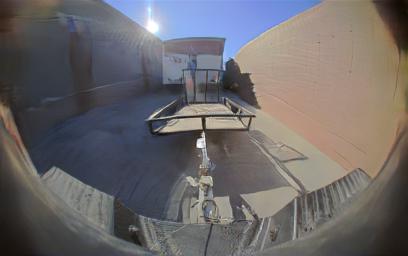}
        \label{fig:result18}
    \end{subfigure}
    \begin{subfigure}[t]{0.16\linewidth}
        \includegraphics[trim={2cm .5cm 3.5cm 0},clip,height=.95\linewidth]{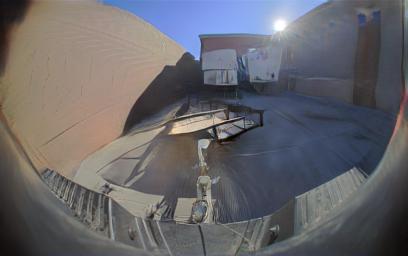}
        \label{fig:result19}
    \end{subfigure}
    \begin{subfigure}[t]{0.16\linewidth}
        \includegraphics[trim={2cm .5cm 3.5cm 0},clip,height=.95\linewidth]{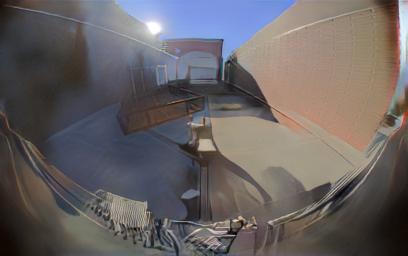}
        \label{fig:result110}
    \end{subfigure}
    \begin{subfigure}[t]{0.16\linewidth}
        \includegraphics[trim={2cm .5cm 3.5cm 0},clip,height=.95\linewidth]{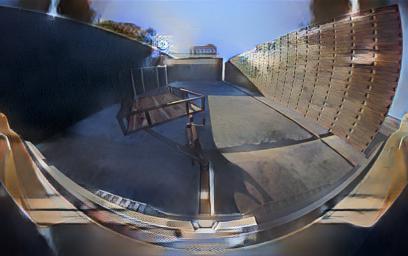}
        \label{fig:result111}
    \end{subfigure}
    \begin{subfigure}[t]{0.16\linewidth}
        \includegraphics[trim={2cm .5cm 3.5cm 0},clip,height=.95\linewidth]{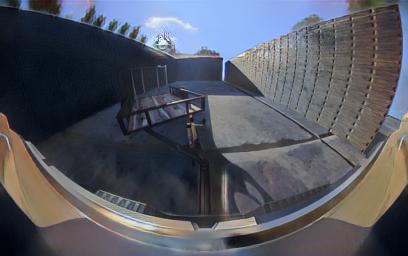}
        \label{fig:result112}
    \end{subfigure}
    \vspace{-0.3cm}
    \begin{subfigure}[t]{0.16\linewidth}
        \includegraphics[trim={2cm .5cm 3.5cm 0},clip,height=.95\linewidth]{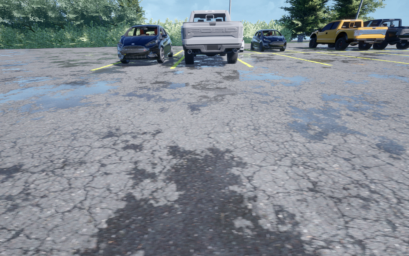}
        \label{fig:result113}
    \end{subfigure}
    \begin{subfigure}[t]{0.16\linewidth}
        \includegraphics[trim={2cm .5cm 3.5cm 0},clip,height=.95\linewidth]{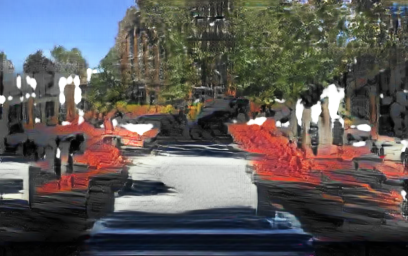}
        \label{fig:result114}
    \end{subfigure}
    \begin{subfigure}[t]{0.16\linewidth}
        \includegraphics[trim={2cm .5cm 3.5cm 0},clip,height=.95\linewidth]{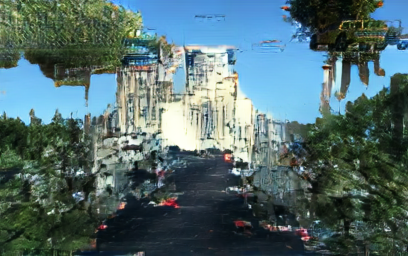}
        \label{fig:result115}
    \end{subfigure}
    \begin{subfigure}[t]{0.16\linewidth}
        \includegraphics[trim={2cm .5cm 3.5cm 0},clip,height=.95\linewidth]{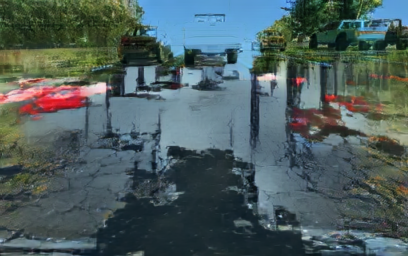}
        \label{fig:result116}
    \end{subfigure}
    \begin{subfigure}[t]{0.16\linewidth}
        \includegraphics[trim={2cm .5cm 3.5cm 0},clip,height=.95\linewidth]{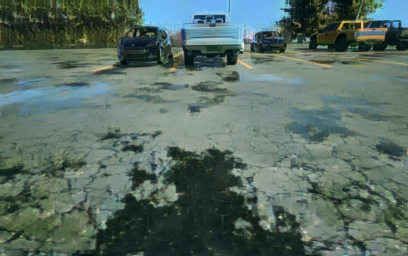}
        \label{fig:result117}
    \end{subfigure}
    \begin{subfigure}[t]{0.16\linewidth}
        \includegraphics[trim={2cm .5cm 3.5cm 0},clip,height=.95\linewidth]{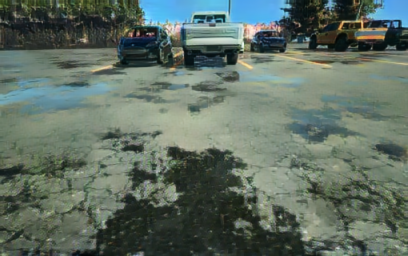}
        \label{fig:result118}
    \end{subfigure}
    \begin{subfigure}[t]{0.16\linewidth}
        \includegraphics[trim={2cm .5cm 3.5cm 0},clip,height=.95\linewidth]{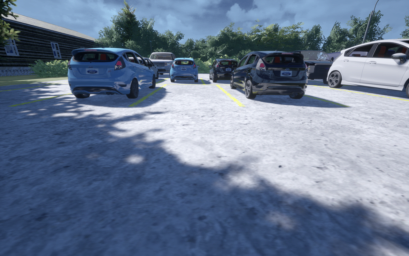}
        \caption{Input}
        \label{fig:result119}
    \end{subfigure}
    \begin{subfigure}[t]{0.16\linewidth}
        \includegraphics[trim={2cm .5cm 3.5cm 0},clip,height=.95\linewidth]{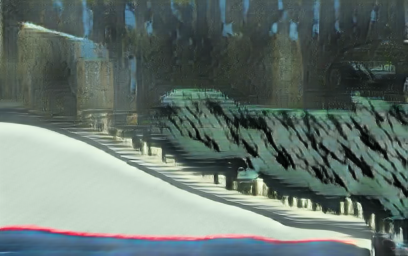}
        \caption{$46\times 46$~\cite{liu2017unsupervised}}
        \label{fig:result120}
    \end{subfigure}
    \begin{subfigure}[t]{0.16\linewidth}
        \includegraphics[trim={2cm .5cm 3.5cm 0},clip,height=.95\linewidth]{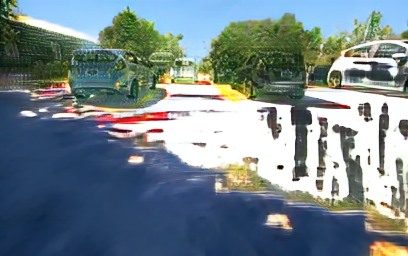}
        \caption{$22 \times 22$}
        \label{fig:result121}
    \end{subfigure}
    \begin{subfigure}[t]{0.16\linewidth}
        \includegraphics[trim={2cm .5cm 3.5cm 0},clip,height=.95\linewidth]{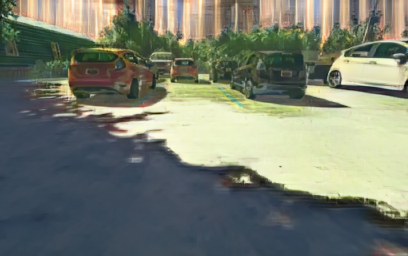}
        \caption{$16 \times 16$}
        \label{fig:result122}
    \end{subfigure}
    \begin{subfigure}[t]{0.16\linewidth}
        \includegraphics[trim={2cm .5cm 3.5cm 0},clip,height=.95\linewidth]{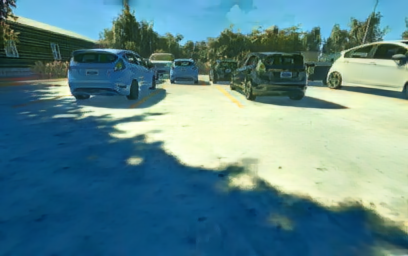}
        \caption{$10 \times 10$}
        \label{fig:result123}
    \end{subfigure}
    \begin{subfigure}[t]{0.16\linewidth}
        \includegraphics[trim={2cm .5cm 3.5cm 0},clip,height=.95\linewidth]{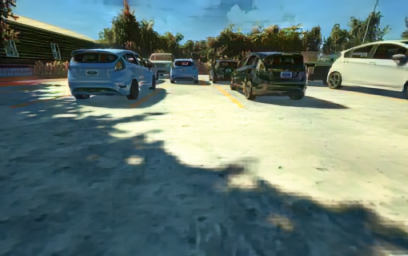}
        \caption{$7 \times 7$}
        \label{fig:result124}
    \end{subfigure}
    \caption{Effect of reducing \texttt{D}'s receptive field on semantic content retention in sim-to-real translations on unpaired and mismatched data: Column (a) shows the input simulated image. Column (b) shows the output of UNIT as-is (with a receptive field of $46\times 46$). Columns (c)-(f) show the sim-to-real outputs of the variants of UNIT with \texttt{D}s with reduced receptive fields. In scenarios where there are multiple models listed with the same receptive field in Fig.~\ref{fig:dismode}, the results shown here are from the model with the most visually appealing results. The top two rows show sim-to-real results on the trailer dataset, while the bottom two rows show results on the parking-highway dataset. Note the gradual improvement in shape, scale and structure of objects present in the scene with a decrease in \texttt{D}'s receptive field, as we move from left to right. Columns (e)-(f) show the best sim-to-real translation results on images in Column (a) in terms of realism and structural coherence.\label{fig:result1}}
\end{figure}

\textbf{Datasets:} The trailer dataset used in this work has $9349$ real images and $9000$ simulated images (from Unreal Engine\footnote{\url{https://en.wikipedia.org/wiki/Unreal_Engine}}) of trailers with straight couplers and A-frame couplers respectively, in daytime on various ground textures. The mismatch in this dataset arises from the difference in trailer coupler shapes between the two domains. We also created another, more conventional dataset that was intentionally mismatched between the two domains. This dataset is called the \emph{parking-highway} dataset. It has $10438$ simulated images of open parking lots in daytime (also from Unreal Engine) and $14172$ real images of daytime highway scenes from BDD100K~\cite{yu2018bdd100k}. Note that the mismatch in the parking-highway dataset arises because of the different background semantics of the simulated and real world images (parking lots vs. highways).

\begin{figure}[h]
    \centering
    \begin{subfigure}[t]{0.16\linewidth}
        \includegraphics[trim={2cm .5cm 3.5cm 0},clip,height=.95\linewidth]{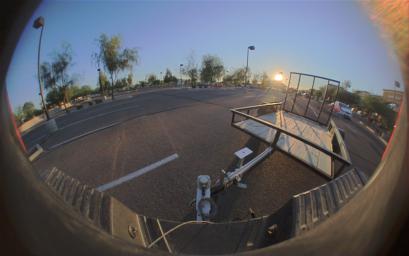}
        \label{fig:result31}
    \end{subfigure}
    \begin{subfigure}[t]{0.16\linewidth}
        \includegraphics[trim={2cm .5cm 3.5cm 0},clip,height=.95\linewidth]{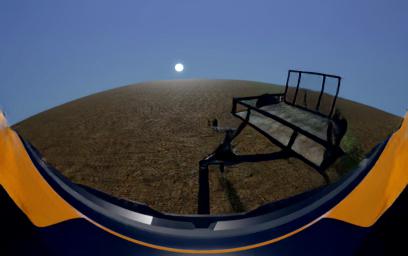}
        \label{fig:result32}
    \end{subfigure}
    \begin{subfigure}[t]{0.16\linewidth}
        \includegraphics[trim={2cm .5cm 3.5cm 0},clip,height=.95\linewidth]{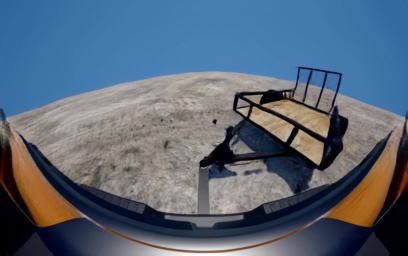}
        \label{fig:result33}
    \end{subfigure}
    \begin{subfigure}[t]{0.16\linewidth}
        \includegraphics[trim={2cm .5cm 3.5cm 0},clip,height=.95\linewidth]{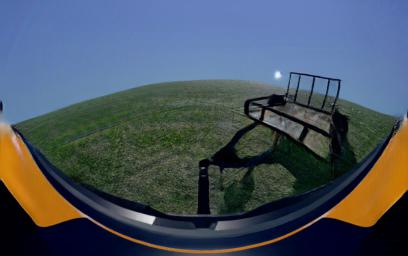}
        \label{fig:result34}
    \end{subfigure}
    \begin{subfigure}[t]{0.16\linewidth}
        \includegraphics[trim={2cm .5cm 3.5cm 0},clip,height=.95\linewidth]{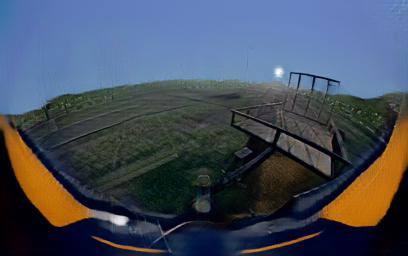}
        \label{fig:result35}
    \end{subfigure}
    \vspace{-0.3cm}
    \begin{subfigure}[t]{0.16\linewidth}
        \includegraphics[trim={2cm .5cm 3.5cm 0},clip,height=.95\linewidth]{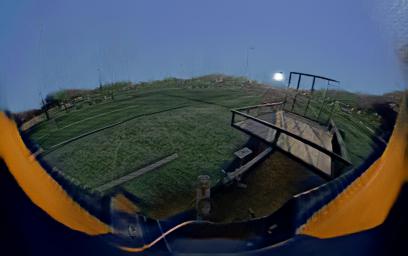}
        \label{fig:result36}
    \end{subfigure}
    \vspace{-0.3cm}
    \begin{subfigure}[t]{0.16\linewidth}
        \includegraphics[trim={2cm .5cm 3.5cm 0},clip,height=.95\linewidth]{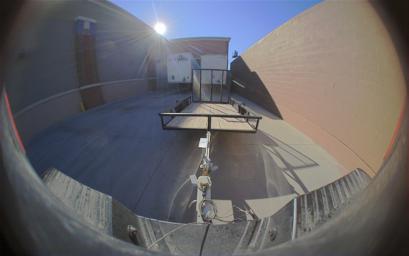}
        \label{fig:result37}
    \end{subfigure}
    \begin{subfigure}[t]{0.16\linewidth}
        \includegraphics[trim={2cm .5cm 3.5cm 0},clip,height=.95\linewidth]{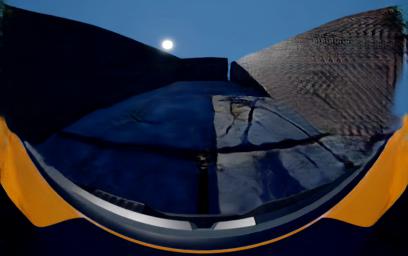}
        \label{fig:result38}
    \end{subfigure}
    \begin{subfigure}[t]{0.16\linewidth}
        \includegraphics[trim={2cm .5cm 3.5cm 0},clip,height=.95\linewidth]{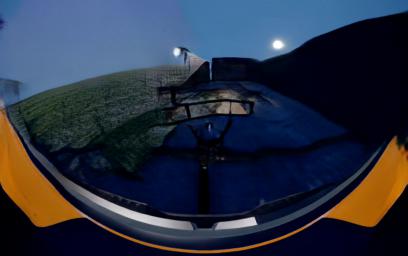}
        \label{fig:result39}
    \end{subfigure}
    \begin{subfigure}[t]{0.16\linewidth}
        \includegraphics[trim={2cm .5cm 3.5cm 0},clip,height=.95\linewidth]{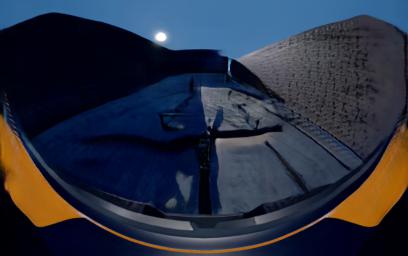}
        \label{fig:result310}
    \end{subfigure}
    \begin{subfigure}[t]{0.16\linewidth}
        \includegraphics[trim={2cm .5cm 3.5cm 0},clip,height=.95\linewidth]{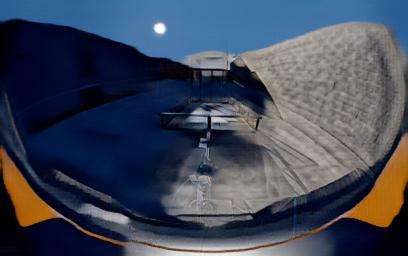}
        \label{fig:result311}
    \end{subfigure}
    \begin{subfigure}[t]{0.16\linewidth}
        \includegraphics[trim={2cm .5cm 3.5cm 0},clip,height=.95\linewidth]{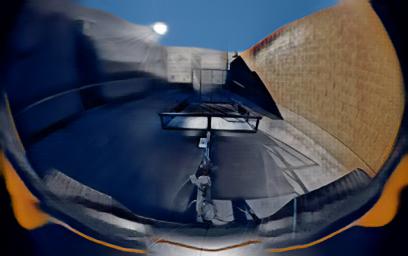}
        \label{fig:result312}
    \end{subfigure}
    \vspace{-0.3cm}
    \begin{subfigure}[t]{0.16\linewidth}
        \includegraphics[trim={2cm .5cm 3.5cm 0},clip,height=.95\linewidth]{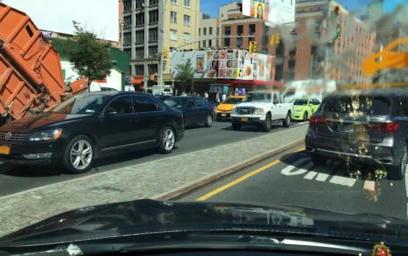}
        \label{fig:result313}
    \end{subfigure}
    \begin{subfigure}[t]{0.16\linewidth}
        \includegraphics[trim={2cm .5cm 3.5cm 0},clip,height=.95\linewidth]{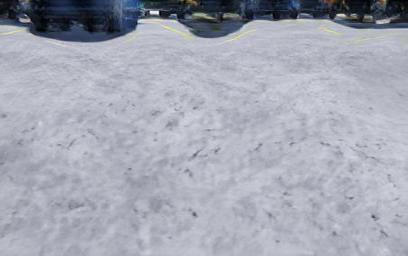}
        \label{fig:result314}
    \end{subfigure}
    \begin{subfigure}[t]{0.16\linewidth}
        \includegraphics[trim={2cm .5cm 3.5cm 0},clip,height=.95\linewidth]{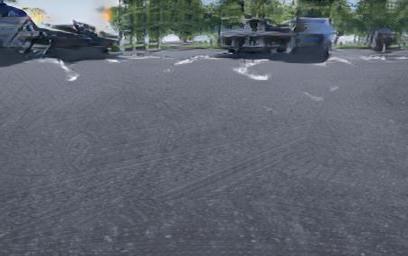}
        \label{fig:result315}
    \end{subfigure}
    \begin{subfigure}[t]{0.16\linewidth}
        \includegraphics[trim={2cm .5cm 3.5cm 0},clip,height=.95\linewidth]{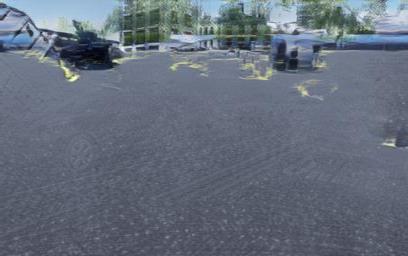}
        \label{fig:result316}
    \end{subfigure}
    \begin{subfigure}[t]{0.16\linewidth}
        \includegraphics[trim={2cm .5cm 3.5cm 0},clip,height=.95\linewidth]{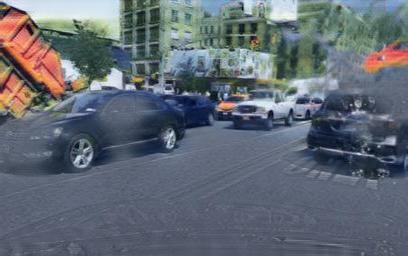}
        \label{fig:result317}
    \end{subfigure}
    \begin{subfigure}[t]{0.16\linewidth}
        \includegraphics[trim={2cm .5cm 3.5cm 0},clip,height=.95\linewidth]{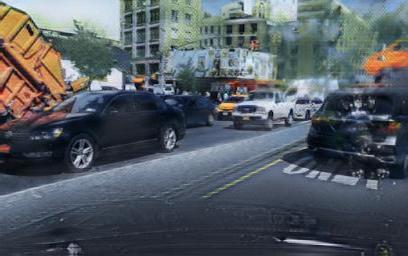}
        \label{fig:result318}
    \end{subfigure}
    \begin{subfigure}[t]{0.16\linewidth}
        \includegraphics[trim={2cm .5cm 3.5cm 0},clip,height=.95\linewidth]{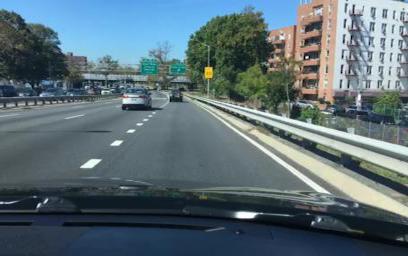}
        \caption{Input}
        \label{fig:result319}
    \end{subfigure}
    \begin{subfigure}[t]{0.16\linewidth}
        \includegraphics[trim={2cm .5cm 3.5cm 0},clip,height=.95\linewidth]{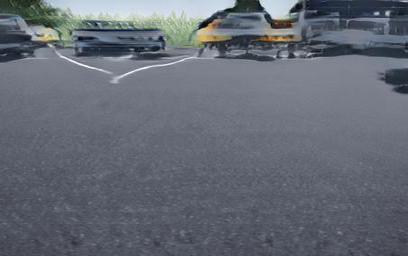}
        \caption{$46\times 46$~\cite{liu2017unsupervised}}
        \label{fig:result320}
    \end{subfigure}
    \begin{subfigure}[t]{0.16\linewidth}
        \includegraphics[trim={2cm .5cm 3.5cm 0},clip,height=.95\linewidth]{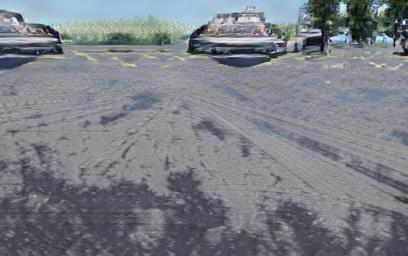}
        \caption{$22 \times 22$}
        \label{fig:result321}
    \end{subfigure}
    \begin{subfigure}[t]{0.16\linewidth}
        \includegraphics[trim={2cm .5cm 3.5cm 0},clip,height=.95\linewidth]{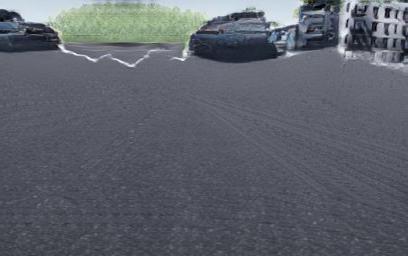}
        \caption{$16 \times 16$}
        \label{fig:result322}
    \end{subfigure}
    \begin{subfigure}[t]{0.16\linewidth}
        \includegraphics[trim={2cm .5cm 3.5cm 0},clip,height=.95\linewidth]{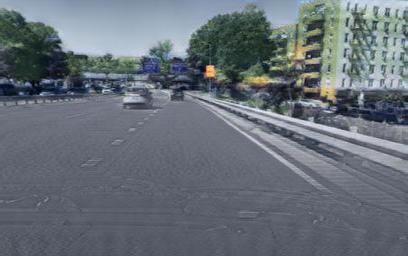}
        \caption{$10 \times 10$}
        \label{fig:result323}
    \end{subfigure}
    \begin{subfigure}[t]{0.16\linewidth}
        \includegraphics[trim={2cm .5cm 3.5cm 0},clip,height=.95\linewidth]{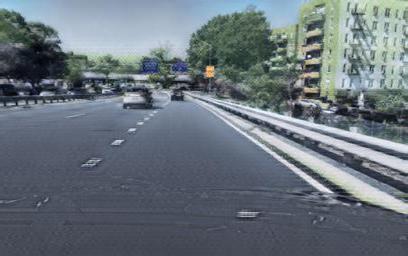}
        \caption{$7 \times 7$}
        \label{fig:result324}
    \end{subfigure}
    \caption{Effect of reducing \texttt{D}'s receptive field on semantic content retention in real-to-sim translations on unpaired and mismatched data: Column (a) shows the input real image. Column (b) shows the output of UNIT as-is (with a receptive field of $46\times 46$). Columns (c)-(f) show the real-to-sim outputs of the variants of UNIT with \texttt{D}s with reduced receptive fields. In scenarios where there are multiple models listed with the same receptive field in Fig.~\ref{fig:dismode}, the results shown here are from the model with the most visually appealing results. The top two rows show real-to-sim results on the trailer dataset, while the bottom two rows show results on the parking-highway dataset. Again, similar to Fig.~\ref{fig:result1}, note the gradual improvement in shape, scale and structure of objects present in the scene with a decrease in \texttt{D}'s receptive field, as we move from left to right. Columns (e)-(f) show the best real-to-sim translation results on images in Column (a) in terms of visual quality and structural coherence.\label{fig:result3}}
\end{figure}

\textbf{Experiment Details:} Our training procedure followed~\cite{liu2017unsupervised} and used ADAM~\cite{kingma2014adam} optimization for training with a learning rate of 0.0001 and momentums of 0.5 and 0.999. Each mini-batch comprises an image from both the domains. All images were resized and then cropped to $256\times 256$ for training. The VGG-16~\cite{simonyan2014very} based perceptual loss was included only in the training of UNIT as-is and not in the training of any of the modified UNIT architectures, as listed in Fig.~\ref{fig:dismode}.

\textbf{Qualitative Results:} Fig.~\ref{fig:result1} shows sim-to-real results for all variants of \texttt{D} listed in Fig.~\ref{fig:dismode}. Since the datasets used for training are mismatched in terms of semantic content, as argued in Section~\ref{sec:intro}, reducing \texttt{D}'s field-of-view (as we move from left to right in Fig.~\ref{fig:result1}), helps generate realistic images while also preserving semantic content. For instance, in the first row, note the gradual decrease in trailer scale and change in shape of the trailer coupler (from straight to A-frame) with a decrease in \texttt{D}'s receptive field (as we move from Column (b) to (f)), to perfectly match the semantic content of the input image in Column (a). In the second row, the input trailer (Column (a)) is masked by shadows. Since most of the real images used for training had a trailer in the image center, the baseline UNIT translates the input image to a realistic looking image with a trailer in the center (Column (b)). Again, as we move from Column (b) to (f), there is an anticlockwise rotation in the orientation of the generated trailer to match the orientation in the input image in Column (a). We observed a similar trend in sim-to-real translations on the parking-highway dataset. In both the third and fourth rows in Fig.~\ref{fig:result1}, the baseline UNIT (Column (b)) masks out the semantic content of the input image (Column (a)), and instead generates a highway road-type scene, as all the real images used for training were from dashcams of vehicles driving on highways. Gradually, as we move from Column (b) to (f), realistic looking parking lot images start getting generated, with semantic content matching that of images in Column (a).

Recall that the baseline UNIT trains sim-to-real and real-to-sim models simultaneously. Fig.~\ref{fig:result3} shows that a reduction in \texttt{D}'s receptive field leads to improved semantic content retention in real-to-sim translations as well. Similar to the results shown in Fig.~\ref{fig:result1}, reducing \texttt{D}'s field-of-view (as we move from left to right in Fig.~\ref{fig:result3}), helps generate images that look synthetic while also preserving semantic content. For instance, in the first row, note the gradual change in shape of the trailer coupler (from A-frame to straight) with a decrease in \texttt{D}'s receptive field (as we move from Column (b) to (f)), to perfectly match the trailer couple shape (straight) in the input image in Column (a). In the second row, the baseline UNIT completely masks out the trailer and replaces it with wall shadows (Column (b)). Again, as we move from Column (b) to (f), we can notice the gradual appearance of a trailer in the same position as that in the input image in Column (a). A similar trend is observed in real-to-sim translations on the parking-highway dataset. In both the third and fourth rows in Fig.~\ref{fig:result3}, the baseline UNIT completely ignores the semantic content of the input images in Column (a) to generate parking lot images, similar to those in the simulated parking data. Again, as we move from Column (b) to (f), highway images that look synthetic start getting generated, with semantic content matching that of images in Column (a).

\textbf{Quantitative Results:} Prior works use metrics such as the IOU over segmentation masks of the input and translated images to quantify content retention~\cite{park2019semantic}. However, none of the open source datasets used for training popular deep segmentation networks (such as Mask R-CNN~\cite{he2017mask}) have a trailer class. Thus, we were unable to use these metrics in this work. The same is true for classification-based metrics, such as the FCN score~\cite{isola2017image}. Therefore, we define a new metric called `Semantic Content Discrepancy' (SCD), which is measured as the Modified Hausdorff Distance (MHD)~\cite{dubuisson1994modified} between the Holistically-nested Edge Detections (HED) of the input simulated image and the sim-to-real translated image (see Fig.~\ref{fig:unit} for example HED outputs). While this metric does have some limitations (refer Appendix for details), in general, a lower SCD indicates that edges, shapes and scales of objects present in the input image are preserved during the sim-to-real translations. This metric was used for quantitative analysis of results from sim-to-real translation of 500 images from the trailer dataset (see Fig.~\ref{fig:mhd}). Note the consistent drop in SCD with a decrease in \texttt{D}'s receptive field (plotted on a log scale). 

\begin{figure}[h]
\centering
    \begin{subfigure}[t]{0.155\linewidth}
        \includegraphics[trim={2cm .5cm 3.5cm 0},clip,height=.95\linewidth]{fig11/original.jpg}
        \label{fig:result21}
    \end{subfigure}
    \begin{subfigure}[t]{0.155\linewidth}
        \includegraphics[trim={2cm .5cm 3.5cm 0},clip,height=.95\linewidth]{fig11/test_vgg.jpg}
        \label{fig:result22}
    \end{subfigure}
    \begin{subfigure}[t]{0.155\linewidth}
        \includegraphics[trim={2cm .5cm 3.5cm 0},clip,height=.95\linewidth]{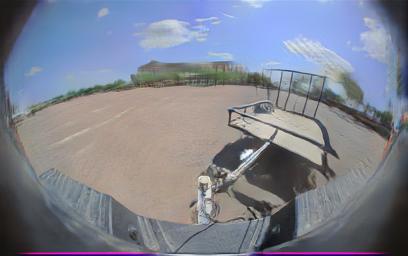}
        \label{fig:result23}
    \end{subfigure}
    ~
    \begin{subfigure}[t]{0.155\linewidth}
        \includegraphics[trim={2cm .5cm 3.5cm 0},clip,height=.95\linewidth]{fig14/original.jpg}
        \label{fig:result24}
    \end{subfigure}
    \begin{subfigure}[t]{0.155\linewidth}
        \includegraphics[trim={2cm .5cm 3.5cm 0},clip,height=.95\linewidth]{fig14/test_vgg.jpg}
        \label{fig:result25}
    \end{subfigure}
        \vspace{-0.3cm}
    \begin{subfigure}[t]{0.155\linewidth}
        \includegraphics[trim={2cm .5cm 3.5cm 0},clip,height=.95\linewidth]{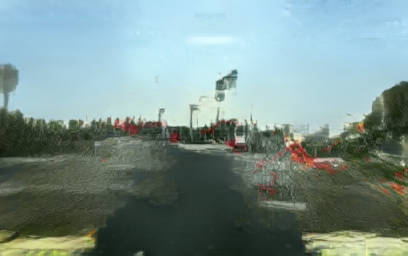}
        \label{fig:result26}
    \end{subfigure}
    \begin{subfigure}[t]{0.155\linewidth}
        \includegraphics[trim={2cm .5cm 3.5cm 0},clip,height=.95\linewidth]{fig12/original.jpg}
        \caption{Simulated trailer}
        \label{fig:result27}
    \end{subfigure}
    \begin{subfigure}[t]{0.155\linewidth}
        \includegraphics[trim={2cm .5cm 3.5cm 0},clip,height=.95\linewidth]{fig12/test_vgg.jpg}
        \caption{UNIT ($46\times 46$)}
        \label{fig:result28}
    \end{subfigure}
    \begin{subfigure}[t]{0.155\linewidth}
        \includegraphics[trim={2cm .5cm 3.5cm 0},clip,height=.95\linewidth]{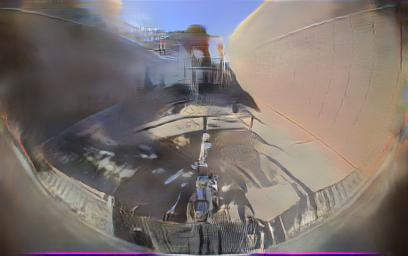}
        \caption{Model 1 ($40\times 40$)}
        \label{fig:result29}
    \end{subfigure}
    ~
    \begin{subfigure}[t]{0.155\linewidth}
        \includegraphics[trim={2cm .5cm 3.5cm 0},clip,height=.95\linewidth]{fig13/original.jpg}
        \caption{Simulated parking}
        \label{fig:result210}
    \end{subfigure}
    \begin{subfigure}[t]{0.155\linewidth}
        \includegraphics[trim={2cm .5cm 3.5cm 0},clip,height=.95\linewidth]{fig13/test_vgg.jpg}
        \caption{UNIT ($46\times 46$)}
        \label{fig:result211}
    \end{subfigure}
    \begin{subfigure}[t]{0.155\linewidth}
        \includegraphics[trim={2cm .5cm 3.5cm 0},clip,height=.95\linewidth]{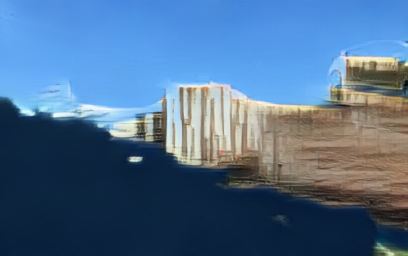}
        \caption{Model 1 ($40\times 40$)}
        \label{fig:result212}
    \end{subfigure}
    \begin{subfigure}[t]{0.155\linewidth}
        \includegraphics[trim={2cm .5cm 3.5cm 0},clip,height=.95\linewidth]{fig11/original.jpg}
        \label{fig:result21}
    \end{subfigure}
    \begin{subfigure}[t]{0.155\linewidth}
        \includegraphics[trim={2cm .5cm 3.5cm 0},clip,height=.95\linewidth]{fig11/magna_modv1_k4.jpg}
        \label{fig:result22}
    \end{subfigure}
    \begin{subfigure}[t]{0.155\linewidth}
        \includegraphics[trim={2cm .5cm 3.5cm 0},clip,height=.95\linewidth]{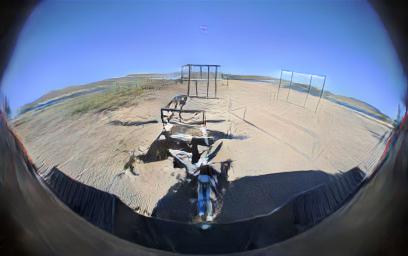}
        \label{fig:result23}
    \end{subfigure}
    ~
    \begin{subfigure}[t]{0.155\linewidth}
        \includegraphics[trim={2cm .5cm 3.5cm 0},clip,height=.95\linewidth]{fig14/original.jpg}
        \label{fig:result21}
    \end{subfigure}
    \begin{subfigure}[t]{0.155\linewidth}
        \includegraphics[trim={2cm .5cm 3.5cm 0},clip,height=.95\linewidth]{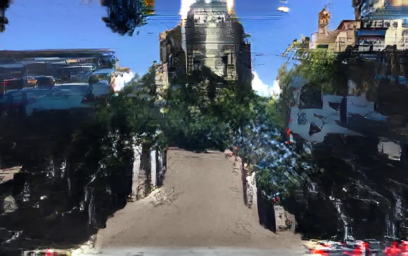}
        \label{fig:result22}
    \end{subfigure}
    \begin{subfigure}[t]{0.155\linewidth}
        \includegraphics[trim={2cm .5cm 3.5cm 0},clip,height=.95\linewidth]{fig14/magna_layer3.jpg}
        \label{fig:result23}
    \end{subfigure}
    \begin{subfigure}[t]{0.155\linewidth}
        \includegraphics[trim={2cm .5cm 3.5cm 0},clip,height=.95\linewidth]{fig12/original.jpg}
        \caption{Simulated trailer}
        \label{fig:result21}
    \end{subfigure}
    \begin{subfigure}[t]{0.155\linewidth}
        \includegraphics[trim={2cm .5cm 3.5cm 0},clip,height=.95\linewidth]{fig12/magna_modv1_k4.jpg}
        \caption{Model 3 ($22\times 22$)}
        \label{fig:result22}
    \end{subfigure}
    \begin{subfigure}[t]{0.155\linewidth}
        \includegraphics[trim={2cm .5cm 3.5cm 0},clip,height=.95\linewidth]{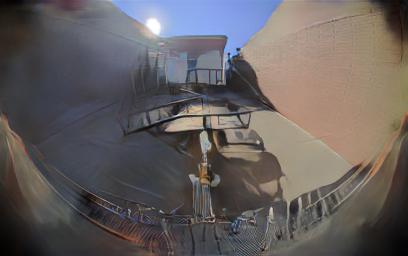}
        \caption{Model 2 ($22\times 22$)}
        \label{fig:result23}
    \end{subfigure}
    ~
    \begin{subfigure}[t]{0.155\linewidth}
        \includegraphics[trim={2cm .5cm 3.5cm 0},clip,height=.95\linewidth]{fig13/original.jpg}
        \caption{Simulated parking}
        \label{fig:result21}
    \end{subfigure}
    \begin{subfigure}[t]{0.155\linewidth}
        \includegraphics[trim={2cm .5cm 3.5cm 0},clip,height=.95\linewidth]{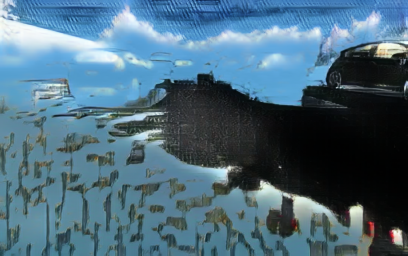}
        \caption{Model 3 ($22\times 22$)}
        \label{fig:result22}
    \end{subfigure}
    \begin{subfigure}[t]{0.155\linewidth}
        \includegraphics[trim={2cm .5cm 3.5cm 0},clip,height=.95\linewidth]{fig13/magna_layer3.jpg}
        \caption{Model 2 ($22\times 22$)}
        \label{fig:result23}
    \end{subfigure}
    \begin{subfigure}[t]{0.155\linewidth}
        \includegraphics[trim={2cm .5cm 3.5cm 0},clip,height=.95\linewidth]{fig11/original.jpg}
        \label{fig:result21}
    \end{subfigure}
    \begin{subfigure}[t]{0.155\linewidth}
        \includegraphics[trim={2cm .5cm 3.5cm 0},clip,height=.95\linewidth]{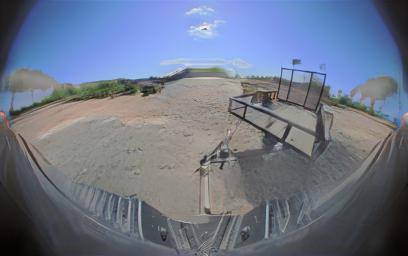}
        \label{fig:result22}
    \end{subfigure}
    \begin{subfigure}[t]{0.155\linewidth}
        \includegraphics[trim={2cm .5cm 3.5cm 0},clip,height=.95\linewidth]{fig11/magna_layer2.jpg}
        \label{fig:result23}
    \end{subfigure}
    ~
    \begin{subfigure}[t]{0.155\linewidth}
        \includegraphics[trim={2cm .5cm 3.5cm 0},clip,height=.95\linewidth]{fig14/original.jpg}
        \label{fig:result21}
    \end{subfigure}
    \begin{subfigure}[t]{0.155\linewidth}
        \includegraphics[trim={2cm .5cm 3.5cm 0},clip,height=.95\linewidth]{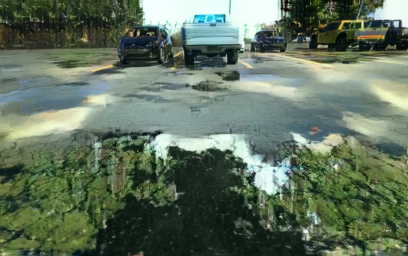}
        \label{fig:result22}
    \end{subfigure}
    \begin{subfigure}[t]{0.155\linewidth}
        \includegraphics[trim={2cm .5cm 3.5cm 0},clip,height=.95\linewidth]{fig14/magna_layer2.jpg}
        \label{fig:result23}
    \end{subfigure}
    \begin{subfigure}[t]{0.155\linewidth}
        \includegraphics[trim={2cm .5cm 3.5cm 0},clip,height=.95\linewidth]{fig12/original.jpg}
        \caption{Simulated trailer}
        \label{fig:result21}
    \end{subfigure}
    \begin{subfigure}[t]{0.155\linewidth}
        \includegraphics[trim={2cm .5cm 3.5cm 0},clip,height=.95\linewidth]{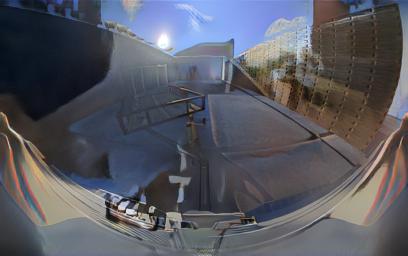}
        \caption{Model 7 ($10\times 10$)}
        \label{fig:result22}
    \end{subfigure}
    \begin{subfigure}[t]{0.155\linewidth}
        \includegraphics[trim={2cm .5cm 3.5cm 0},clip,height=.95\linewidth]{fig12/magna_layer2.jpg}
        \caption{Model 5 ($10\times 10$)}
        \label{fig:result23}
    \end{subfigure}
    ~
    \begin{subfigure}[t]{0.155\linewidth}
        \includegraphics[trim={2cm .5cm 3.5cm 0},clip,height=.95\linewidth]{fig13/original.jpg}
        \caption{Simulated parking}
        \label{fig:result21}
    \end{subfigure}
    \begin{subfigure}[t]{0.155\linewidth}
        \includegraphics[trim={2cm .5cm 3.5cm 0},clip,height=.95\linewidth]{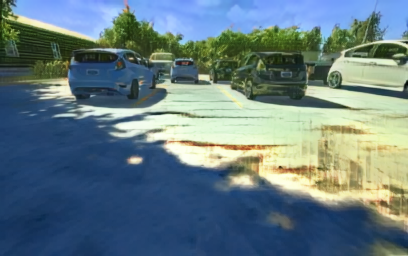}
        \caption{Model 7 ($10\times 10$)}
        \label{fig:result22}
    \end{subfigure}
    \begin{subfigure}[t]{0.155\linewidth}
        \includegraphics[trim={2cm .5cm 3.5cm 0},clip,height=.95\linewidth]{fig13/magna_layer2.jpg}
        \caption{Model 5 ($10\times 10$)}
        \label{fig:result23}
    \end{subfigure}
    \caption{\emph{Role of overfitting:} Effect of reducing the number of parameters in \texttt{D} while simultaneously maintaining its receptive field on sim-to-real translation with mismatched data. Columns (a), (b), (c); (g), (h), (i) and (m), (n), (o) show results on the trailer dataset, while Columns (d), (e) and (f); (j), (k), (l) and (p), (q), (r) show results on the parking-highway dataset. Columns (a), (d), (g), (j), (m) and (p) show input simulated images. The rest of the columns show sim-to-real translations results, for the simulated images to their left, with the models listed in the column captions.\label{fig:result2}}
\end{figure}

\textbf{Role of Overfitting:} Fig.~\ref{fig:result2} shows that reducing the number of parameters/layers in \texttt{D} (which in effect also reduces its representation power and consequently reduces the generator's chances of overfitting) while keeping its receptive field the same as the baseline UNIT architecture, does not improve structural coherence in sim-to-real translations (see Columns (a)-(f)). A quantitative analysis of the sim-to-real translation results on 500 simulated trailer images gives an average SCD of 12.20 with Model 1 (4 layers), as compared to that of 11.03 with UNIT (5 layers). Thus, simply reducing the number of parameters in the discriminator does not result in improved semantic content retention for a receptive field of $46 \times 46$. The same trend is observed with discriminator models with a receptive field of $22\times 22$ in all scenarios except for the second example from the parking-highway dataset (see Columns (g)-(l)). Again, a quantitative analysis of the sim-to-real translation results on 500 simulated trailer images gives an average SCD of 11.21 with Model 2 (4 layers), as compared to that of 10.35 with Model 3 (5 layers). However, as shown in Columns (m)-(r), reducing the representation power of the discriminator in the case of a receptive field of $10\times 10$ improves semantic content retention during translation as we see better results with Model 5 (3 layers) as compared to those with Model 7 (4 layers). Quantitatively also Model 5 gives an average SCD of 9.51 as compared to that of 11.06 with Model 7. All these results combined lead us to conclude that reducing the number of parameters in the discriminator architecture while keeping its receptive field the same does not always lead to better results. However, for discriminator architectures with a small enough receptive field ($10\times 10$ in this case), designing an architecture with fewer parameters could lead to a lower SCD, while maintaining the desired level of photorealism (again, see Columns (m)-(r)).  

%%%%%% DISCUSSION %%%%%%
\section{Discussion}
\label{sec:discussion}
\textbf{Realism vs. Semantic Content Retention:} Fig.~\ref{fig:result1} and Fig.~\ref{fig:result3} establish the fact that reducing D's field of view, i.e. receptive field, which is equivalent to weakening the realism signal during GAN training, helps reduce SCD in unsupervised image translation. However, we would like to note that this SCD improvement comes at the cost of photorealism, specifically in the case of the trailer dataset. For instance, in the first row in Fig.~\ref{fig:result1}, sim-to-real translation using the baseline UNIT architecture (with a receptive field of $46\times 46$) gives a more photorealistic image than that using Model 8 (with a receptive field of $7\times 7$). Thus, there exists a two-way tradeoff between photorealism and semantic content retention during translation. Extensive experimentation with \texttt{D}'s architecture is required to find the sweet spot of receptive field that simultaneously achieves desired levels of photorealism and semantic content retention during translation.

\textbf{Limitations of the SCD metric:} Recall that SCD is defined as the MHD between the Holistically-nested Edge Detections (HED) of the input simulated image and the sim-to-real translated image. Thus, in effect, SCD quantifies the difference between edge maps of the input and translated images. This is beneficial with respect to object entities, such as trailers and cars, for which simulation provides ground truth labels in the form of bounding boxes. However, a difference between edge masks also penalizes visual artifacts generated during translation, such as trees and building in the background (see Fig.~\ref{fig:unit12}). The generation of such artifcats might infact be desirable for certain perception tasks, such as trailer detection, as such artifacts add diversity to the training images. Thus, whether or not SCD is the right metric to quantify which sim-to-real model does best in terms of both realism and semantic content retention is very much dependent on the downstream perception task.

%%%%%% CONCLUSION & FUTURE WORK %%%%%%
\section{Conclusion}
\label{sec:conclusion}
Experiments with \texttt{D}'s architecture in UNIT~\cite{liu2017unsupervised} on two mismatched datasets show that \texttt{D}'s receptive field is directly correlated with semantic content discrepancy of the generated image in sim-to-real image translation. Thus, reducing \texttt{D}'s receptive field reduces semantic content discrepancy during translation, as shown both qualitatively and quantitatively in Section~\ref{sec:results}. However, to effectively learn sim-to-real translation, the two datasets cannot be completely mismatched. For e.g., one cannot expect to learn what real trees look like if the real images are of cars only. Furthermore, diversity in the training datasets from the two domains, regardless of the content mismatch between them, is still essential as GANs very quickly `memorize' and thus, are highly susceptible to overfitting.

\subsubsection*{Acknowledgments}
The authors would like to sincerely thank Rohan Bhasin, Gautham Sholingar and Xianling Zhang for generating the synthetic data used in this work. We are also indebted to Gaurav Pandey for insightful discussions; Jinesh Jain, Marcos P. Gerardo-Castro and Sandhya Sridhar for reviewing the submission draft. Finally, we would like to express our immense gratitude to Raju Nallapa and Ken Washington for their continuous support and encouragement.

\bibliography{ref}
\end{document}